\documentclass{article}

     \usepackage[nonatbib,preprint]{neurips_2019}

\usepackage[utf8]{inputenc} 
\usepackage[T1]{fontenc}    
\usepackage{hyperref}       
\usepackage{url}            
\usepackage{booktabs}       
\usepackage{amsfonts}       
\usepackage{nicefrac}       
\usepackage{microtype}      

\usepackage{graphicx}
\usepackage{subfig}
\usepackage{caption}

\usepackage{amsmath,amsgen,amstext,amssymb}
\usepackage{amsfonts, amsthm}
\usepackage{dsfont}

\DeclareMathAlphabet{\mathpzc}{OT1}{pzc}{m}{it}
\DeclareMathOperator*{\argmax}{\arg\max}

\newtheorem{theorem}{Theorem}[section]

\newtheorem{definition}{Definition}[section]

\newcommand{\cM}{{\cal M}}

\newcommand{\bbA}{{\mathbb{A}}}

\newcommand{\bbQ}{{\mathbb{Q}}}

\newcommand{\bbV}{{\mathbb{V}}}
\newcommand{\bbX}{{\mathbb{X}}}

\newcommand{\var}{\mathbb{V}\mbox{ar}}

\newcommand{\ex}{\mathbb{E}}
\newcommand{\pr}{\mathbb{P}}

\newcommand{\beqn}{\begin{equation}}
\newcommand{\eeqn}{\end{equation}}

\newcommand{\cF}{{\mathcal{F}}}

\newcommand{\cT}{{\mathcal{T}}}

\newcommand\1{\mathds{1}}
\newcommand{\indi}[1]{\1_{\{#1\}}}

\newcommand\Reals{{\mathbb{R}}}
\newcommand\Ints{{\mathbb{Z}}}

\newcommand{\setA}{{\bbA}}
\newcommand{\setQ}{{\bbQ}}
\newcommand{\setV}{{\bbV}}
\newcommand{\setX}{{\bbX}}
\newcommand{\rewardR}{{R}}
\newcommand{\BellmanOp}{{\cT_{B}}}
\newcommand{\ConsistentOp}{{\cT_{C}}}
\newcommand{\GenericOp}{{\cT}}
\newcommand{\RobustOp}{{\cT_{\beta_k}}}
\newcommand{\RobustOpHat}{{\cT_{\hat{\beta}_k}}}
\newcommand{\RobustOpTilde}{{\cT_{\tilde{\beta}_k}}}
\newcommand{\RobustOpFamily}{{\cT_{\beta}^\cF}}

\title{A Family of Robust Stochastic Operators for Reinforcement Learning}

\author{%
  Yingdong Lu, Mark S.~Squillante, Chai W.~Wu \\
  Mathematical Sciences\\
  IBM Research\\
  Yorktown Heights, NY 20198, USA\\
  \texttt{\{yingdong, mss, cwwu\}@us.ibm.com} \\
}

\begin{document}

\maketitle

\begin{abstract}
We consider a new family of stochastic operators for reinforcement learning that seek to alleviate negative effects and become more
robust to approximation or estimation errors.
Theoretical results are established, showing that our family of operators preserve optimality and increase the action gap in a stochastic sense.
Empirical results illustrate the strong benefits of our robust stochastic operators, significantly outperforming the classical Bellman
and recently proposed operators.
\end{abstract}

\section{Introduction}
\label{sec:intro}
Reinforcement learning has a rich history within the machine learning community to solve a wide variety of decision making problems
in environments with unknown and possibly unstructured dynamics.
Through iterative application of a convergent operator, value-based reinforcement learning (RL) generates successive refinements of an initial
value function
\cite{KaLiMo96,Szep10,SutBar11}.
$Q$-learning \cite{Watk89} is a particular RL technique in which the value iteration computations consist of evaluating the
corresponding Bellman equation
without a model of the environment.

While $Q$-learning continues to be broadly and successfully used in RL to determine the optimal actions of an agent, the development of new $Q$-learning approaches
that improve convergence speed, accuracy and robustness remains of great interest.
One area of particular interest concerns environments in which there exist approximation or estimation errors.
Of course, when no approximation/estimation errors are present, then the corresponding Markov Decision Process (MDP) can be solved exactly with the Bellman operator.
However, in the presence of nonstationary errors~--~a typically encountered example being when a discrete-state, discrete-time MDP is used to approximate a
continuous-state, continuous-time system~--~then the optimal state-action value function obtained through the Bellman operator does not always describe the value of
stationary policies.
When the differences between the optimal state-action value function and the suboptimal state-action value functions are small, this can result in errors in identifying
the truly optimal actions.

To help explain and formalize this phenomena, Farahmand~\cite{Fara11} introduced the notion of action-gap regularity and showed that a larger action-gap regularity
implies a smaller loss in performance.
Building on action-gap regularity and its benefits with respect to (w.r.t.) performance loss, Bellemare et al.~\cite{Bellemare:2016} considered a particular approach
to having the value iteration converge to an alternative action-value function $Q$ associated with the same optimal action policy~--~i.e., maintain properties of
optimality-preserving~--~while at the same time achieving a larger separation between the $Q$-values of optimal actions and those of suboptimal actions~--~i.e.,
maintain properties of action-gap increasing.
The former properties ensure optimality whereas the latter properties assist the value-iteration algorithm to determine the optimal actions of an agent faster,
more easily, and with less errors of mislabeling suboptimal actions.
Hence, by exploiting weaker optimality conditions than the Bellman equation and due to the known benefits of increasing the action gap, this approach can lead to
alternatives to the classical Bellman operator that improve the convergence speed, accuracy and robustness of RL in environments with approximation/estimation errors,

Following this approach, Bellemare et al.~\cite{Bellemare:2016} propose deterministic operator alternatives to the classical Bellman operator and show that the
proposed operators satisfy the properties of optimality-preserving and gap-increasing.
Then, after empirically demonstrating the benefits of their proposed deterministic operator alternatives, the authors raise a number of open fundamental questions
w.r.t.\ the possibility of weaker conditions for optimality, the statistical efficiency of their proposed operators, and the possibility of finding a
maximally efficient operator.

At the heart of the problem of interest is a fundamental trade-off between the degree to which the preservation of optimality is violated and the degree to which
the action gap is increased.
Although the benefits of increasing the action gap in the presence of approximation/estimation errors are known \cite{Fara11}, increasing the action gap beyond a
certain region in a deterministic sense can lead to violations of optimality preservation (as a result of deviating too far from Bellman optimality), thus rendering
value iterations that may not ensure convergence to optimal solutions.
Hence, any deterministic operator alternative is unfortunately quite limited in the degree to which it can be both gap-increasing and optimality-preserving, and
thus in turn quite limited in the degree to which it can address the problems of approximation/estimation errors in RL.

We therefore consider an approach based on a novel stochastic framework that can increase the action gap well beyond such a deterministic region for individual
value iterations, via a random variable (r.v.), while controlling in a probabilistic manner the overall value iterations, via a sequence of r.v.s, to ensure
optimality preservation in a stochastic sense.
Our general approach is applicable to arbitrary $Q$-value approximation schemes in which the sequence of r.v.s provide support to devalue suboptimal actions while
preserving the set of optimal policies almost surely (a.s.), thus making it possible to increase the action gap between the $Q$-values of optimal and suboptimal
actions beyond the deterministic region;
this can be critically important in practice because of the known advantages of increasing the action gap when there are approximation/estimation errors.
In devising a family of operators within our framework endowed with these properties, we provide a general stochastic approach that addresses the inherent
deficiencies of deterministic operator alternatives to the classical Bellman operator and that yields greater robustness w.r.t.\ mislabeling suboptimal
actions in the presence of approximation/estimation errors.
To the best of our knowledge, this paper presents the first proposal and theoretical analysis of such types of general robust stochastic operators, which is an
approach not often seen in the RL literature and should be exploited to a much greater extent.

The research literature contains a wide variety of studies of operator alternatives to the Bellman operator, including the $\epsilon$-greedy method~\cite{Watk89},
speedy $Q$-learning~\cite{AzMuGa+11}, policy iteration-like $Q$-learning~\cite{BerYu12}, and the Boltzmann softmax operator and its variants~\cite{AsaLit17}.
Each of these operator alternatives seek to address certain issues in RL.
In this paper we complement these previous studies of operator alternatives and focus on operators that seek to achieve greater robustness w.r.t.\
approximation/estimation errors; in fact, our empirical studies are based on $Q$-learning with the $\epsilon$-greedy method.

Our theoretical results include proving
that our stochastic operators are optimality-preserving and gap-increasing in a stochastic sense.
Since the value-iteration sequence generated under our stochastic operators is based on realizations of r.v.s, our family of robust stochastic operators subsumes
the family of deterministic operators in \cite{Bellemare:2016} as a strict subset (because the realizations of r.v.s can be fixed to match that of the deterministic
operators as a special case).
We further prove that stochastic and variability orderings among the sequence of random operators lead to corresponding orderings among the action gaps,
implying that greater variability in the sequence of r.v.s renders a larger action gap.
Our stochastic framework and theoretical results shed new light on the open fundamental questions raised in \cite{Bellemare:2016}, which includes our family of
robust stochastic operators rendering significantly weaker conditions for optimality.
Another important implication of our results is that the search space for the maximally efficient operator should be an infinite dimensional space of r.v.s,
instead of the finite space alluded to in \cite{Bellemare:2016}.
Yet another important implication
is that our stochastic ordering results provide order relationships among the sequence of random operators
w.r.t.\ action gaps,
which can in turn lead to determining the best sequence of r.v.s and can even lead to maximally efficient operators.
These results further extend our understanding of the relationship between action-gap regularity and the effectiveness of $Q$-learning in environments with
approximation/estimation errors beyond the initial studies in \cite{Fara11,Bellemare:2016}.
Our proofs of these theoretical results involve mathematical arguments and important technical details that are unique to stochastic operators and stochastic
ordering, and distinct from any previous instances of deterministic operators.
For technical details on convergence of probability measures, stochastic ordering and related stochastic results underlying our theoretical results,
we refer the interested reader to \cite{Billingsley99,ChoTei03,shaked2007stochastic}.

We subsequently apply our robust stochastic operators to obtain empirical results for various problems in the OpenAI Gym framework \cite{AIgym}.
Using these existing codes with minor modifications, we compare the empirical results under our family of stochastic operators against those under both the
classical Bellman operator and the consistent Bellman operator from \cite{Bellemare:2016}.
These experiments consistently show that our robust stochastic operators outperform both
of these deterministic operators due to a larger action gap.
Appendix~\ref{app:code} of the supplement provides the corresponding python code modifications used in our experiments, which we will release on GitHub and
make publically available.

\section{Preliminaries}
We consider a standard RL framework (see, e.g., \cite{BerTsi96}) in which a learning agent interacts with a stochastic environment.
This interaction is modeled as a discrete-space, discrete-time discounted
MDP
given by $(\setX, \setA, \pr, \rewardR, \gamma)$,
where $\setX$ is the set of states, $\setA$ the set of actions, $\pr$ the transition probability kernel, $\rewardR$ the reward function mapping state-action pairs
to a bounded subset of $\Reals$, and $\gamma \in [0,1)$ the discount factor.
Let $\setQ$ and $\setV$ denote the set of bounded real-valued functions over $\setX \times \setA$ and $\setX$, respectively.
For $Q \in \setQ$, define $V(x) := \max_a Q(x,a)$ and use the same definition for variants such as $\hat{Q} \in \setQ$ and $\hat{V}(x)$.
Let $x'$ always denotes the next state r.v.
For the current state $x$ in which action $a$ is taken, i.e., $(x,a) \in \setX \times \setA$,
denote by $\pr(\cdot | x,a)$ the conditional transition probability for the next state $x'$ and
define $\ex_{\pr} := \ex_{x' \sim \pr(\cdot | x,a)}$ to be the expectation w.r.t.\ $\pr(\cdot | x,a)$.

A stationary policy $\pi(\cdot | x) : \setX \rightarrow \setA$ defines the distribution of control actions given the current state $x$, which reduces to
a deterministic policy when the conditional distribution renders a constant action for each state $x$;
with slight abuse of notation, we always write the policy $\pi(x)$.
The stationary policy $\pi$ induces a value function $V^\pi : \setX \rightarrow \Reals$ and an action-value function
$Q^\pi : \setX \times \setA \rightarrow \Reals$ where $V^\pi(x) := Q^\pi(x,\pi(x))$ defines the expected discounted cumulative
reward under policy $\pi$ starting in state $x$ and where $Q^\pi$ satisfies the Bellman equation
\begin{equation}
Q^\pi(x,a) = \rewardR(x,a) + \gamma \ex_{\pr} Q^\pi(x',\pi(x')).
\label{eq:Bellman}
\end{equation}
Our goal is to determine a policy $\pi^*$ that achieves the optimal value function $V^*(x) := \sup_\pi V^\pi(x), \forall x \in \setX$,
which also produces the optimal action-value function $Q^*(x,a) := \sup_\pi Q^\pi(x,a), \forall (x,a) \in \setX \times \setA$.
The \emph{Bellman operator} $\BellmanOp : \setQ \rightarrow \setQ$ is defined pointwise as
\begin{equation}
\BellmanOp Q(x,a) := \rewardR(x,a) + \gamma \ex_{\pr} \max_{b \in \setA} Q(x',b) ,
\label{op:Bellman}
\end{equation}
or equivalently
$\BellmanOp Q(x,a) = \rewardR(x,a) + \gamma \ex_{\pr} V(x')$.
The Bellman operator $\BellmanOp$ is known (see, e.g., \cite{BerTsi96}) to be a contraction mapping in supremum norm,
and its unique fixed point coincides with the optimal action-value function, namely
$Q^*(x,a) = \rewardR(x,a) + \gamma \ex_{\pr} \max_{b \in \setA} Q^*(x',b)$,
or equivalently
$Q^*(x,a) = \rewardR(x,a) + \gamma \ex_{\pr} V^*(x')$.
This in turn indicates that the optimal policy $\pi^*$ can be obtained by
$\pi^*(x) = \argmax_{a \in \setA} Q^*(x,a), \; \forall x \in \setX$.

While the Bellman operator can exactly solve the MDP when there are no approximation/estimation errors, the previously noted differences
between optimal and suboptimal state-action value functions in the presence of such errors can result in wrongly identifying the optimal actions.
To address these and related nonstationary effects of approximation/estimation errors arising in practice,
Bellemare et al.~\cite{Bellemare:2016} propose the so-called \emph{consistent Bellman operator} defined as
\begin{equation}
\ConsistentOp Q(x,a) := \rewardR(x,a) + \gamma \ex_{\pr} [ \indi{x\neq x'} \max_{b \in \setA} Q(x',b) + \indi{x=x'} Q(x,a) ],
\label{op:Consistent}
\end{equation}
where $\indi{\cdot}$ denotes the indicator function.
The consistent Bellman operator $\ConsistentOp$ preserves a local form of stationarity by redefining the action-value function $Q$ such that,
if an action $a \in \setA$ is taken from the state $x \in \setX$ and the next state $x'=x$, then action $a$ is taken again.
Bellemare et al.~\cite{Bellemare:2016}
proceed to show that the consistent Bellman operator yields the optimal policy $\pi^*$,
and in particular that $\ConsistentOp$ is both optimality-preserving and gap-increasing,
according to (deterministic) definitions that they provide which are compatible with those from Farahmand~\cite{Fara11}.

\section{Robust Stochastic Operators}
\label{sec:robust}
In this section we present our stochastic framework which includes proposing a general family of robust stochastic operators,
providing precise definitions of the concepts of optimality-preserving and gap-increasing in a stochastic sense for a sequence of random operators, 
and establishing that any sequence of this general family of operators is optimality-preserving and gap-increasing.
Our introduction of a new family of operators and our shifting the focus from one deterministic operator to a sequence of stochastic operators has
significant implications w.r.t.\ the open questions raised in \cite{Bellemare:2016}.
Specifically, our results show that the conditions for optimality are much weaker and the statistical efficiency of our operators can be made much stronger,
both allowing significant degrees of freedom in finding alternatives to the Bellman operator for different purposes and applications.
Meanwhile, these important improvements completely alter and clarify the question of finding the maximally efficient operators from a finite dimensional
parameter optimization problem suggested in \cite{Bellemare:2016} to an optimization problem in an infinite dimensional space (of the infinite sequence of r.v.s),
for which we establish that a stochastic ordering among the sequence of random operators leads to a corresponding ordering among the action gaps.
It is important to note that our approach can be extended to variants of the Bellman operator such as SARSA \cite{RumNir94}, policy evaluation \cite{Sutt88}
and fitted $Q$-iteration \cite{ErGeWe05}.

For all $Q_0 \in \setQ$, $x \in \setX$, $a \in \setA$ and sequences $\{ \beta_k : k\in \Ints_+ \}$ of independent nonnegative r.v.s with finite support and
expectation $\overline{\beta}_k := \ex_{\beta}[\beta_k] \in [0,1]$, we define
\begin{equation}
\RobustOp Q_k(x,a) := \rewardR(x,a) + \gamma \ex_{\pr} \max_{b \in \setA} Q_k(x',b) - \beta_k (V_k(x) - Q_k(x,a)),
\label{op:Robust}
\end{equation}
or equivalently
$\RobustOp Q_k(x,a) := \rewardR(x,a) + \gamma \ex_{\pr} V(x') - \beta_k (V_k(x) - Q_k(x,a))$.
(Note that the operator in \eqref{op:Robust} is equivalent to the Bellman operator whenever the action $a$ is optimal or $\beta_k=0$, thus making the difference
term $0$ in these cases.)
Then members of the general family of robust stochastic operators include the sequence $\{ \RobustOp \}$ defined over all probability distributions
for each r.v.\ in the sequence $\{ \beta_k \}$ with $\overline{\beta}_k \in [0,1]$.
(Note, in particular, that the r.v.s $\beta_k$ can follow a different probability distribution for each $k$.)
We further define $\RobustOpFamily$ to be the general family of robust stochastic operators comprising all sequences of operators $\{ \GenericOp \}$,
each as defined in~\eqref{op:Robust}, such that there exists a sequence of $\{\beta_k\}$ and, for all $x \in \setX$ and $a \in \setA$, the following
inequalities hold
\begin{equation*}
\BellmanOp Q(x,a) - \beta_k (V_k(x) - Q_k(x,a)) \le \GenericOp Q(x,a) \le \BellmanOp Q(x,a).
\end{equation*}
Obviously, these conditions are strictly weaker than those identified in \cite{Bellemare:2016}
(the former based on r.v.s $\beta_k$ that can take on values outside of $[0,1)$, the latter being deterministic and constrained to $[0,1)$).
Since the r.v.s $\beta_k$ need not be identically distributed (with the sole requirement that $\overline{\beta}_k$ is between $0$ and $1$ inclusively) and
since realizations of $\beta_k$ can take on values far beyond or equal to $1$, the family of operators $\RobustOpFamily$ clearly subsumes the family of
previously identified deterministic operators.

For the analysis of our family of stochastic operators, we consider the following key definitions.
%
\begin{definition}
A sequence of random operators $\{\GenericOp_k\}$ for $\cM = (\setX, \setA, \pr, \rewardR, \gamma)$ is \emph{optimality-preserving in a stochastic sense}
if for any $Q_0 \in \setQ$ and $x \in \setX$, and for the sequence of r.v.s $Q_{k+1} := \GenericOp_k Q_k$, the following properties hold:
$V_k(x):= \max_{a \in \setA} Q_k(x,a)$ converges almost surely (a.s.) to a constant $\hat{V}(x)$ as $k\rightarrow \infty$,
and for all $a \in \setA$, we have a.s.
\begin{equation}
Q^*(x,a) < V^*(x) \Rightarrow \limsup_{k\rightarrow\infty} Q_k(x,a) < \hat{V}(x) .
\label{eq:opt-pre}
\end{equation}
\end{definition}
\begin{definition}
A sequence of random operators $\{\GenericOp_k\}$ for $\cM = (\setX, \setA, \pr, \rewardR, \gamma)$ is \emph{gap-increasing in a stochastic sense} if for all
$Q_0 \in \setQ$, $x \in \setX$ and $a \in \setA$, the following inequality holds a.s.:
\begin{equation}
A(x,a) := \liminf_{k\rightarrow\infty} [V_k(x) - Q_k(x,a)] \geq V^*(x) - Q^*(x,a) .
\label{eq:gap-inc}
\end{equation}
\end{definition}
%

The property of the optimality-preserving definition
essentially ensures a.s.\ that at least one optimal action remains optimal and all suboptimal actions remain suboptimal.
At the same time,
the property of the gap-increasing definition
implies robustness when the inequality \eqref{eq:gap-inc} is strict a.s.\ for at least one $(x,a) \in \setX \times \setA$.
In particular, as the action gap of an operator increases while remaining optimality-preserving, the end result is greater robustness to
approximation/estimation errors \cite{Fara11}.

We next present one of our main theoretical results establishing that our general family of robust stochastic operators is both optimality-preserving
and gap-increasing in a stochastic sense.
\begin{theorem} \label{thm:main}
Let $\BellmanOp$ be the Bellman operator defined in \eqref{op:Bellman} and
$\{ \RobustOp \}$ a sequence of robust stochastic operators as defined in \eqref{op:Robust}.
Considering the sequence of r.v.s $Q_{k+1} := \RobustOp Q_k$ on a sample path basis with $Q_0 \in \setQ$,
the sequence of operators $\{ \RobustOp \}$ is both optimality-preserving and gap-increasing in a stochastic sense, a.s.
Furthermore, all operators in the family $\RobustOpFamily$ are optimality-preserving and gap-increasing in a stochastic sense, a.s.
\end{theorem}

Even though the stochastic operators in $\RobustOpFamily$ are not contraction mappings and therefore do not have a fixed point (as true also for
$\ConsistentOp$~\cite{Bellemare:2016}), Theorem~\ref{thm:main} establishes that each of these stochastic operators in $\RobustOpFamily$ are optimality-preserving.
Moreover, the definition of $\RobustOpFamily$ and Theorem~\ref{thm:main} significantly enlarge the set of optimality-preserving and gap-increasing deterministic
operators identified in \cite{Bellemare:2016}.
In particular, our new sufficient conditions for optimality-preserving operators in a stochastic sense implies that significant deviation from the Bellman operator
is possible without loss of optimality;
cf., the deterministic operator in~\cite{Bellemare:2016} never allows a value of $\beta_k$ equal to or greater than $1$.
More importantly, the definition of $\RobustOpFamily$ and Theorem \ref{thm:main} imply that the search space for maximally efficient operators is
an infinite dimensional space of r.v.s, instead of the finite dimensional space for maximally efficient operators alluded to in \cite{Bellemare:2016}.
We now establish results on stochastic ordering properties among the sequences of r.v.s $\{ \beta_k \}$ that lead to corresponding ordering properties among the
action gaps of the random operators; these results offer key relational insights into important orderings of different operators in $\RobustOpFamily$
such as greater variability in the sequence of r.v.s rendering a larger action gap.
This can then be exploited in searching for and attempting to find maximally efficient operators in practice.

\begin{theorem}
Suppose $\hat{Q}_{k+1}$ and $\tilde{Q}_{k+1}$ are respectively updated with two different robust stochastic operators $\RobustOpHat$ and $\RobustOpTilde$
that are distinguished by $\hat{\beta}_k$ and $\tilde{\beta}_k$ satisfying $\hat{\beta}_k \ge_{st}\tilde{\beta}_k$;
namely $\hat{Q}_{k+1}= \RobustOpHat \hat{Q}_k$ and $\tilde{Q}_{k+1}= \RobustOpTilde \tilde{Q}_k$, where $\ge_{st}$ denotes the stochastic ordering.
Then we have that the action gaps of the two systems are stochastically ordered in the same direction, namely $\hat{A}(x,a) \ge_{st} \tilde{A}(x,a)$.
\label{thm:var1}
\end{theorem}

\begin{theorem}
Suppose $\hat{Q}_{k+1}$ and $\tilde{Q}_{k+1}$ are respectively updated with two different robust stochastic operators $\RobustOpHat$ and $\RobustOpTilde$
that are distinguished by $\hat{\beta}_k$ and $\tilde{\beta}_k$ satisfying $\hat{\beta}_k \ge_{cx}\tilde{\beta}_k$;
namely $\hat{Q}_{k+1}= \RobustOpHat \hat{Q}_k$ and $\tilde{Q}_{k+1}= \RobustOpTilde \tilde{Q}_k$, where $\ge_{cx}$ denotes the convex ordering.
Then we have that the action gaps of the two systems are convex ordered in the same direction, namely $\hat{A}(x,a) \ge_{cx} \tilde{A}(x,a)$.
\label{thm:var2}
\end{theorem}

\begin{theorem}
Suppose $\hat{Q}_{k+1}$ and $\tilde{Q}_{k+1}$ are respectively updated with two different robust stochastic operators $\RobustOpHat$ and $\RobustOpTilde$
that are distinguished by $\hat{\beta}_k$ and $\tilde{\beta}_k$ satisfying $\ex[\hat{\beta}_k] = \ex[\tilde{\beta}_k]$ and $\var[\hat{\beta}_k] \le \var[\tilde{\beta}_k]$;
namely $\hat{Q}_{k+1}= \RobustOpHat \hat{Q}_k$ and $\tilde{Q}_{k+1}= \RobustOpTilde \tilde{Q}_k$.
Then we have $\var[\hat{Q}_{k+1}] \le \var[\tilde{Q}_{k+1}]$.
\label{thm:var3}
\end{theorem}

The first two theorems conclude that, among the sequences of $\beta_k$ that preserve optimality, those stochastically larger and more variable sequences
can produce larger action gaps.
Theorem \ref{thm:var3} points out that a larger variance for $\beta_k$, with the same fixed mean value, leads to a larger variance for $Q_k(x,a)$.
We know that, in the limit, the optimal action will maintain its state-action value function.
Then, when $k$ is sufficiently large, we can expect that the state-value function $Q_k(x,b^*)$ for the optimal action $b^*$ in state $x$ will be very close to the
optimal value $Q^*(x,b^*)$.
A larger variance therefore implies that there is a higher likelihood that the state-value function $Q_k(x,a)$ for sub-optimal actions $a$ will take on smaller values,
and thus they can be understood to have a larger action gap.
Hence, using $\beta_k$ with large variances can be seen as a simple implementation of the stochastic ordering results.  
Finally, we note that the results of Theorem~\ref{thm:var3} are consistent with our observations from the numerical experiments in Section~\ref{sec:experiments} where
the operator $\RobustOp$ associated with the sequence $\{ \beta_k \}$ drawn from a uniform distribution outperforms the operator $\RobustOpHat$ associated with the
constant sequence $\{ \beta_k = \overline{\beta} \}$.

\section{Experimental Results}
\label{sec:experiments}
Within the general RL framework of interest, we consider a standard, yet generic, form for $Q$-learning so as to cover the various problems empirically
examined in this section.
Specifically, for all $Q_0 \in \setQ$, $x \in \setX$, $a \in \setA$ and an operator of interest $\GenericOp$, we consider the sequence of action-value
$Q$-functions based on the following generic update rule:
\begin{equation}
Q_{k+1}(x,a) = (1-\alpha_k) Q_k(x,a) + \alpha_k \GenericOp Q_k(x,a) ,
\label{eq:RL}
\end{equation}
where $\alpha_k$ is the learning rate for iteration $k$.
Our empirical comparisons consist of the Bellman operator $\BellmanOp$, the consistent Bellman operator $\ConsistentOp$,
and instances of our family of robust stochastic operators $\RobustOp$, denoted throughout this section as RSO.

We conduct various experiments across several well-known problems using the OpenAI Gym framework \cite{AIgym}, namely
Acrobot, Mountain Car, Cart Pole and Lunar Lander.
This collection of problems spans a variety of RL examples with different characteristics, dimensions, parameters, and so on.
In each case, the state space is continuous and discretized to a finite set of states;
namely, each dimension is discretized into equally spaced bins where
the number of bins depends on the problem to be solved and the reference codebase that is used.
For every problem, the specific Q-learning algorithms considered are defined as in \eqref{eq:RL} where the appropriate operator of interest
$\BellmanOp$, $\ConsistentOp$ or $\RobustOp$ is substituted for $\GenericOp$;
at each timestep, \eqref{eq:RL} is applied to a single point of the $Q$-function at the current state and action.
The experiments for each problem from the OpenAI Gym were run using the existing code found at \cite{OpenAIcode1,OpenAIcode2} exactly as is
with the \emph{sole} change consisting of the replacement of the Bellman operator in the code with corresponding implementations of either
the consistent Bellman operator or RSO; see Appendix~\ref{app:code} of the supplement for the corresponding python code.
It is apparent from these codes that RSO can be directly and easily implemented as a replacement for the classical Bellman operator.
We note that each of these algorithms from the OpenAI Gym implements a form of the $\epsilon$-greedy method
(e.g., occasionally picking a random action or using a randomly perturbed $Q$-function to determine the action)
to enable some form of exploration in addition to the exploitation-based search of the optimal policy using the $Q$-function.

Multiple experimental trials are run for each problem, where we ensured the setting of the random starting state to be the same in
each experimental trial for all three types of operators by initializing them with the same random seed.
We observe across all experimental results that for different problems and different variants of the Q-learning algorithm, simply replacing
the Bellman operator or the consistent Bellman operator with the RSO generally results in significant performance improvements.
The RSO considered in every set of experimental trials for each problem consists of different distributions for $\beta_k$.
We initially present in what immediately follows the RSO results for $\beta_k$ uniformly distributed over $[0,2)$,
i.e., $\beta_k \sim U [0,2)$, and then consider the RSO results under a different distribution for $\beta_k$.
Note that a significant number of experiments were performed with different combinations of distributions for $\beta_k$ over the iterations
$k\in \Ints_+$~--~e.g., $\beta_k \sim U[0,1)$ for $\beta_0, \ldots, \beta_{k'}$ and then $\beta_k \sim U[0,2)$ for $\beta_{k'+1}, \ldots$~--~but
these results were not considerably better, and often worse, than those presented below for $\beta_k \sim U [0,2)$.
We also find that the relative performance of these operators did not depend significantly on the amount of exploration of the $\epsilon$-greedy algorithm;
in particular, the same trends were observed over a wide range of values for $\epsilon$.


\begin{figure}[htbp]
\centering
\subfloat[Acrobot problem.\label{fig:acrobot1}]{\includegraphics[width=0.525\textwidth]{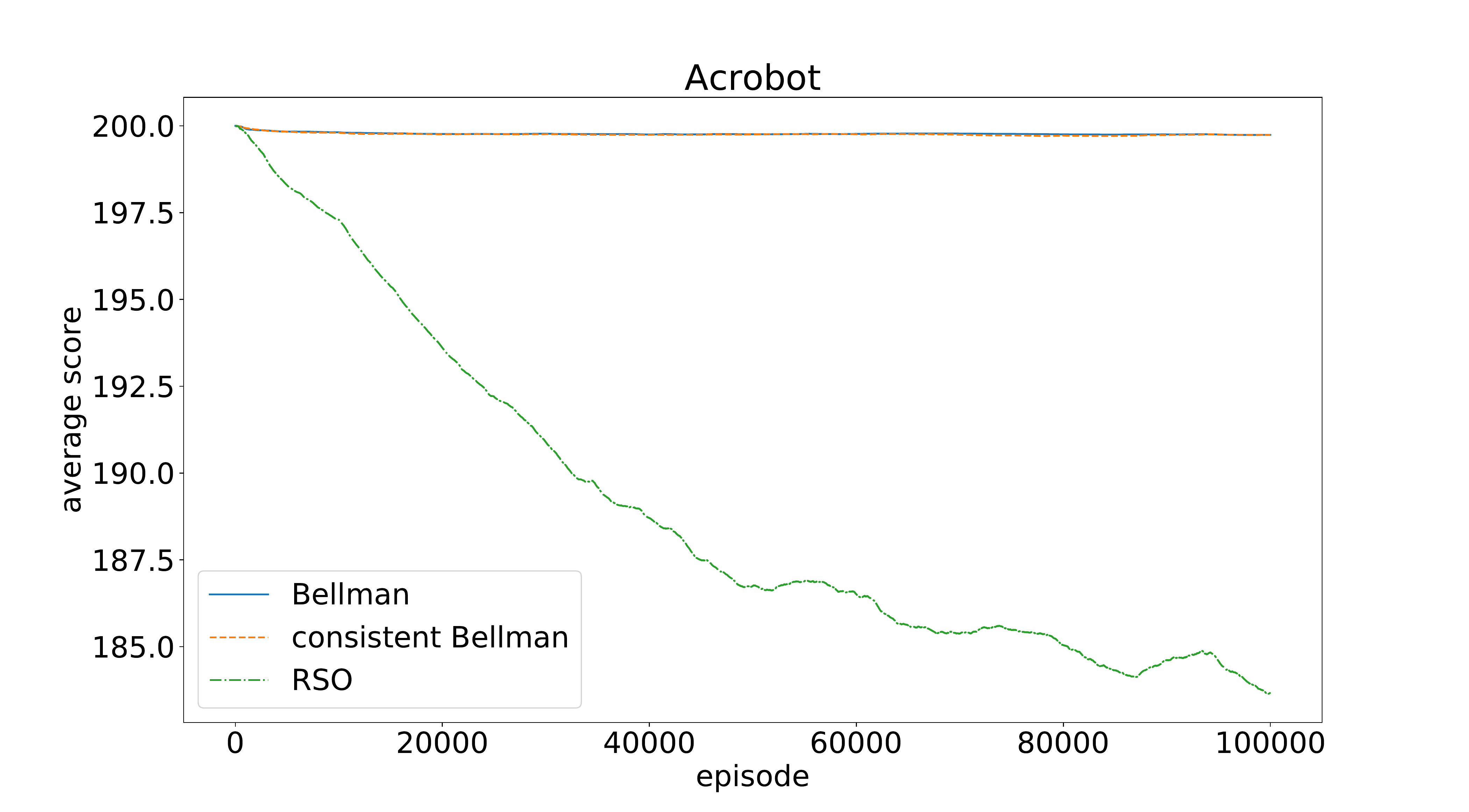}}
\subfloat[Mountain Car problem.\label{fig:mountaincar1}]{\includegraphics[width=0.525\textwidth]{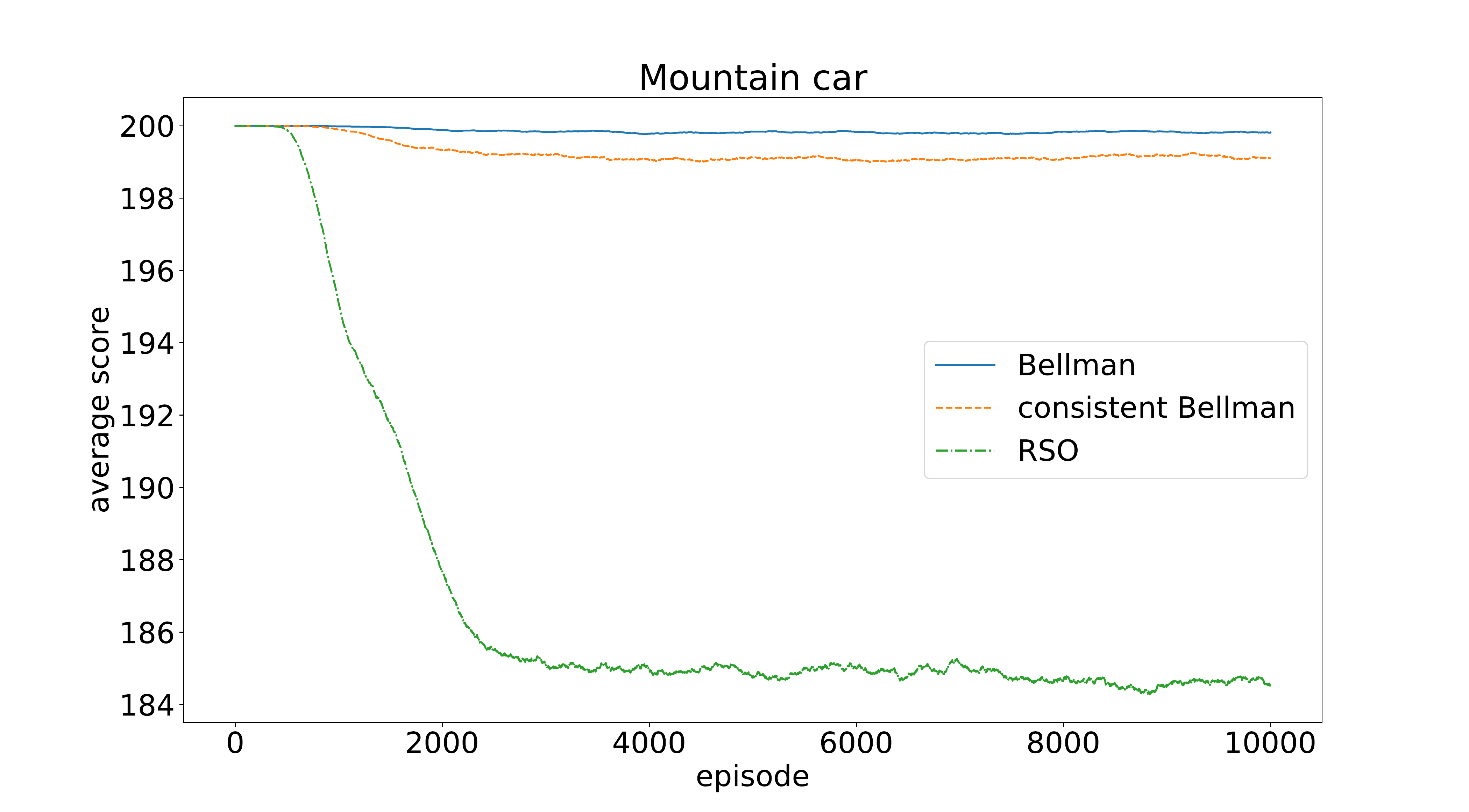}}
\caption{Average number of steps needed to solve minimization problems during training.}
\label{fig:acrobot-mountaincar}
\end{figure}

\subsection{Acrobot}
This problem is first discussed in \cite{acrobot}.
The state vector is $6$-dimensional with three actions possible in each state,
and the score represents the number of timesteps needed to solve the problem.
The position and velocity are discretized into $8$ bins whereas the other state components are discretized into $10$ bins.
We ran $20$ experimental trials over many episodes, with a goal of \emph{minimizing} the score.

Figure~\ref{fig:acrobot1} plots the score, averaged over moving windows of $1000$ episodes across the $20$ trials, as a function of the number of episodes.
We observe that the average score under the RSO exhibits much better performance than under the Bellman operator or the consistent Bellman operator.
Furthermore, as can be seen from the smoothness of the curves in Figure~\ref{fig:acrobot1}, the standard deviation is relatively small for all three operators.

A primary objective of our stochastic operator framework is to realize performance improvements in the presence of approximation/estimation errors
through increasing the action gap.
As a representative example to illustrate and quantify the larger action gap achieved under the RSO, we run each of the three algorithms for $100,000$ episodes
and consider the mean action gap in the last episode, averaged over $40$ trials.
This experiment shows that the Bellman operator, Consistent Bellman operator and RSO has a mean action gap of 0.1436, 0.1263 and 0.9004, respectively, thus
quantifying the significantly larger action gap realized under the RSO.
%


\subsection{Mountain Car}
This problem is first discussed in \cite{mountaincar}.
The state vector is $2$-dimensional with a total of three possible actions,
and the score represents the number of timesteps needed to solve the problem.
The state space is discretized into a $40\times 40$ grid.
We ran $20$ experimental trials over $10,000$ episodes for training,
each of which consists of up to $200$ steps;
then the problem is solved for $1000$ episodes using the policy obtained from the Q-function training.
In both cases, the goal is to \emph{minimize} the score.

For the training phase, Figure~\ref{fig:mountaincar1} plots the score, averaged over moving windows of $500$ episodes across the $20$ trials,
as a function of the number of episodes.
We observe that the average score under the RSO exhibits much better performance than under the Bellman operator or the consistent Bellman operator.
Moreover, as can be seen from the smoothness of the curves in Figure~\ref{fig:mountaincar1}, the standard deviation is relatively small for all three operators.
For the testing phase,
the average score and the standard deviation of the score over the $20$ experimental trials, each comprising $1000$ episodes, are respectively given by:
$129.93$ and $32.68$ for the Bellman operator;
$127.58$ and $30.90$ for the consistent Bellman operator; and
$122.70$ and $7.25$ for the RSO.
Here we observe that both the average score and its standard deviation under the RSO exhibit better performance than under the Bellman operator
or the consistent Bellman operator.



\subsection{Cart Pole}
This problem is first discussed in \cite{cartpole}.
The state vector is $4$-dimensional with two actions possible in each state,
and the score represents the number of steps where the cart pole
stays upright before either falling over or going out of bounds.
The position and velocity are discretized into $8$ bins whereas the angle and angular velocity are discretized into $10$ bins.
%
%
We ran $20$ experimental trials over many episodes, each of which consists
of up to $200$ steps with a goal of \emph{maximizing} the score.
When the score is above 195, the problem is considered solved.

We compute
the score, averaged over moving windows of $1000$ episodes across the $20$ trials, as a function of the number of episodes.
Similarly,
the corresponding standard deviation
is computed,
taken over the same number of score values.
We observe that both the average score and its standard deviation under the RSO exhibit better performance than under the Bellman operator
or the consistent Bellman operator.
Specifically,
the average score over the last $100$ episodes across the $20$ trials is $190.92$ under the RSO
in comparison with $183.67$ and $184.07$ under the Bellman and consistent Bellman operators, respectively;
the corresponding standard deviations of the scores are $19.44$, $28.87$ and $27.82$ for the RSO, Bellman operator and consistent operator, respectively.
%



\begin{figure}[htbp]
\centering
\subfloat[Average score during training phase.\label{fig:lunar1}]{\includegraphics[width=0.525\textwidth]{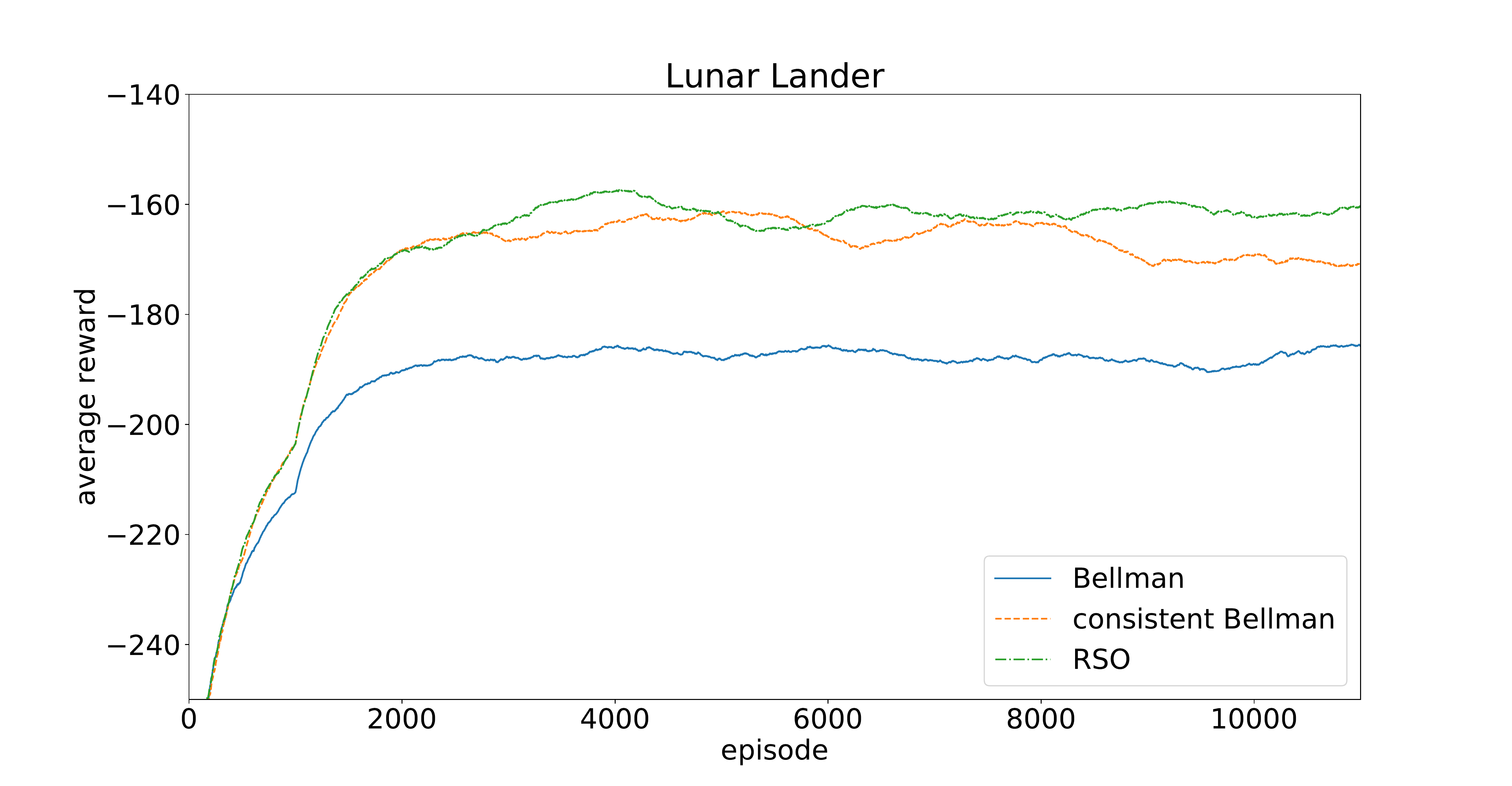}}
\subfloat[Distribution of scores during testing phase.\label{fig:lunar2}]{\includegraphics[width=0.525\textwidth]{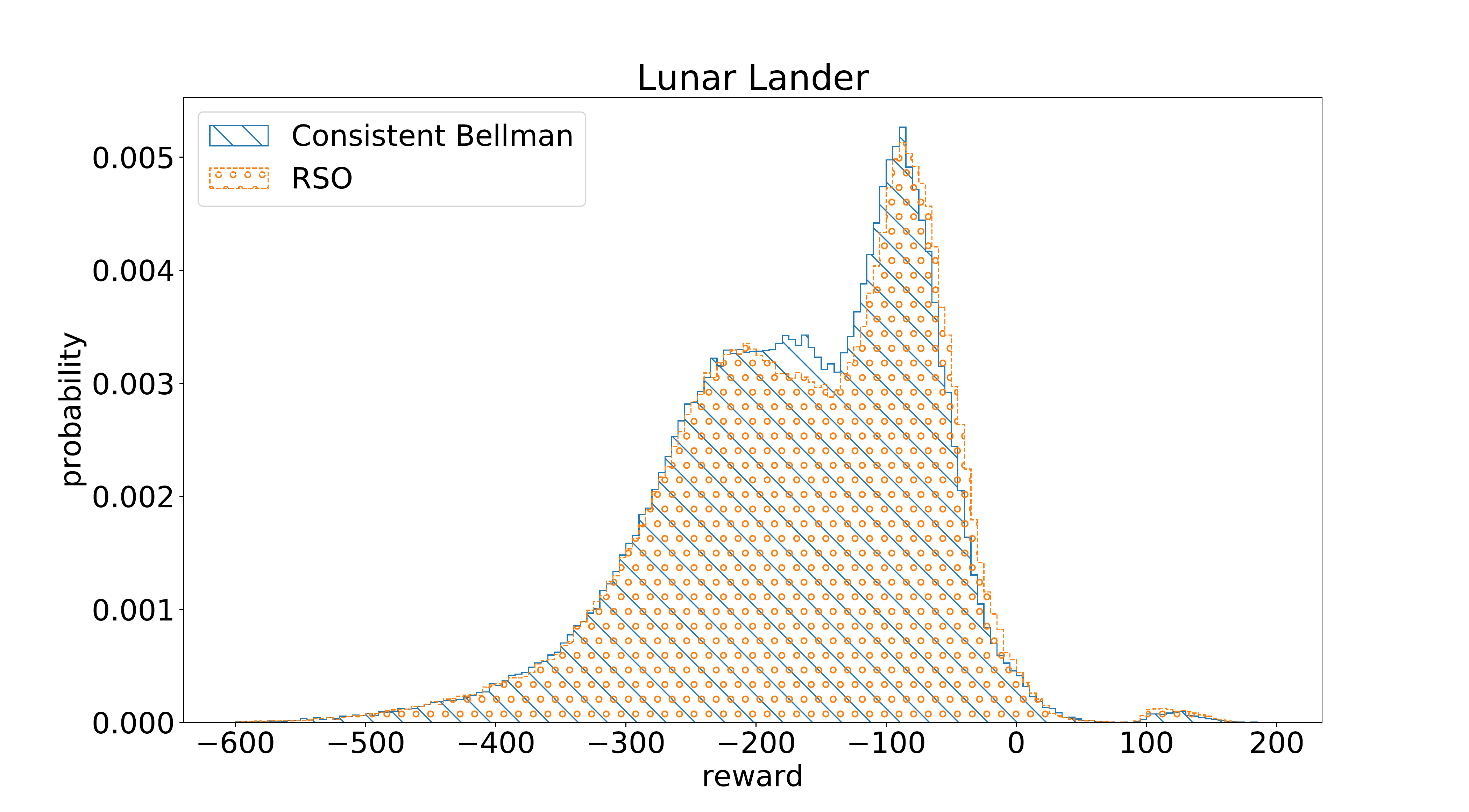}}
\caption{Statistics on score in solving Lunar Lander maximization problem.}
\label{fig:lunar}
\end{figure}


\subsection{Lunar Lander}
This problem is discussed in \cite{AIgym}.
The state vector is $8$-dimensional with a total of four possible actions,
and the physics of the problem is known to be notoriously more difficult than the foregoing problems.
The $6$ continuous state variables are discretized into $4$ bins each.
The score represents the cumulative reward comprising positive points for successful degrees of landing and negative points for fuel usage and crashing.
We ran $20$ experimental trials over many episodes, each of which consists
of up to $200$ steps with a goal of \emph{maximizing} the score.

For the training phase, Figure~\ref{fig:lunar1} plots the score, averaged over moving windows of $1000$ episodes across the $20$ trials,
as a function of the number of episodes.
We observe that the average score under the RSO exhibits better performance than under the Bellman operator or the consistent Bellman operator.
Moreover, as can be seen from the smoothness of the curves in Figure~\ref{fig:lunar1}, the standard deviation is relatively small for all three operators.
For the testing phase, the average score over the $20$ experimental trials, each comprising $1000$ episodes, is respectively given by:
$-241.94$ for the Bellman operator;
$-188.44$ for the consistent Bellman operator; and
$-167.51$ for the RSO.
(Once again, the standard deviation is comparable across all three operators.)
Here we observe yet again that the average score under the RSO exhibits better performance than under the Bellman operator or the consistent Bellman operator.
The improved performance under the RSO can be explained by Figure~\ref{fig:lunar2} that plots the distribution of scores for both the RSO and the consistent Bellman operator.
Here we observe that significant portions of the distribution for the RSO are shifted further to the right.

\subsection{Instances of the RSO}
Lastly, we consider the performance of the RSO under different probability distributions for the sequence of r.v.s $\{ \beta_k \}$.
The performance results for the RSO in the above subsections were based on $\beta_k \sim U [0,2)$.
We also compare these results against the corresponding performance results for the RSO with $\beta_k \sim U [0,1)$ and $\beta_k = 1, \forall k$.
These results clearly demonstrate that the RSO with $\beta_k \sim U[0,2)$ tends to consistently performs better than the RSO with either $\beta_k \sim U[0,1)$
or fixed $\beta_k=1.0$ across the various problems from the OpenAI Gym, consistent with our theoretical results.

\section{Conclusions}
\label{sec:conclusions}
Building on the work of Farahmand~\cite{Fara11} and Bellemare et al.~\cite{Bellemare:2016},
who show that increasing the action gap while preserving optimality can improve the performance of value-iteration algorithms in environments with approximation/estimation errors,
we propose and analyze a new general family of robust stochastic operators for reinforcement learning that subsumes as a strict subset the classical Bellman operator and other
deterministic operators proposed in the literature.
Our theoretical results include proving that our stochastic operators are optimality-preserving and gap-increasing in a stochastic sense and that stochastic and variability
orderings among the sequence of random operators result in corresponding orderings among the action gaps, the latter implying that greater variability in the sequence leads
to a larger action gap.
In addition to improved robustness w.r.t.\ approximation/estimation errors, our theoretical results shed new light on the open fundamental questions raised in \cite{Bellemare:2016}
related to weaker optimality conditions, statistical efficiency of proposed deterministic operators, and finding maximally efficient operators.
A collection of empirical
results~--~based on well-known problems within the OpenAI Gym framework spanning various reinforcement learning examples with diverse characteristics~--~consistently
demonstrates and quantifies the significant performance improvements obtained with our operators over existing operators.
Our family of robust stochastic operators represents an approach not often seen in the reinforcement learning literature that should be exploited to a much greater extent.

%
%


\bibliography{main}
\bibliographystyle{abbrv}

\appendix
%
%
%

\clearpage
\newpage
\newpage
\newpage

\section{Proofs of Theoretical Results}

\subsection{Proof of Theorem~\ref{thm:main}}
%
For any $x \in \setX$, define $\pi^{b_k}(x):=\argmax_bQ_k(x,b)$ on each sample path $\omega$ of the stochastic operator $\{ \RobustOp \}$.
By the definition of $Q_k(x,a)$, we have that $\RobustOp Q_k(x,a) \leq \BellmanOp Q_k(x,a)$ and $\RobustOp Q_k(x,\pi^{b_k}(x)) = \BellmanOp Q_k(x,\pi^{b_k}(x))$,
both a.s.
Although $\pi^{b_k}$ depends on the state $x$, we will omit the argument $x$ in what follows for ease of exposition.

Making use of the facts that  $Q_{k}(x, \pi^{b_k})=\RobustOp  Q_{k-1}(x,\pi^{b_k})= V_k(x)$, we then derive
\begin{align*}
V_{k+1}(x) - V_k(x) & \ge Q_{k+1}(x, \pi^{b_k}) -  Q_{k}(x, \pi^{b_k}) \\
& = \RobustOp Q_k(x, \pi^{b_k}) -\RobustOp  Q_{k-1}(x,\pi^{b_k}) =\BellmanOp Q_k(x,\pi^{b_k})-\RobustOp  Q_{k-1}(x,\pi^{b_k}) \\
& =\BellmanOp Q_{k-1}(x,\pi^{b_k}) + \gamma \ex_{\pr}[V_{k}(x')-V_{k-1}(x')|x, \pi^{b_k}] -\RobustOp  Q_{k-1}(x,\pi^{b_k})\\
& \ge \gamma\ex_{\pr}[V_{k}(x')-V_{k-1}(x')|x, \pi^{b_k}] .
 \end{align*}
This renders
\begin{align*}
V_{k+1}(x)- V_k(x) &\ge \gamma\ex_{\pr}[V_{k}(x')-V_{k-1}(x')|x, \pi^{b_k}],
\end{align*}
and by induction we obtain
\begin{align*}
V_{k+1}(x)- V_k(x) &\ge \gamma^k \ex_{\pr}[V_{1}(x')-V_{0}(x')|x, \pi^{b_1},\ldots, \pi^{b_k}],
\end{align*}
from which we conclude 
\begin{align}
\label{eqn:geo_tail}
V_{k+1}(x)- V_k(x) &\ge -\gamma^k ||V_{1}(x')-V_{0}(x')||_\infty.
\end{align}
Define $f_k= ||V_{1}(x')-V_{0}(x')||_\infty\sum_{\ell=0}^{k-1} \gamma^\ell$ for $k\in \Ints^+$.
Then \eqref{eqn:geo_tail} implies that $V_k(x) + f_k$ is monotone, and thus it will converge.
Meanwhile, $f_k$ obviously converges to $||V_{1}(x')-V_{0}(x')||_\infty/(1-\gamma)$, which leads to the a.s.\ convergence of $V_k(x)$. 

Given the a.s.\ convergence of $V_k(x)$, we now need to identify its limit.
The probabilistic nature of the stochastic operators makes it possible to leverage different forms of convergence of measures for the
corresponding sequence of r.v.s.
Specifically, it suffices for us to establish the limit of $V_k(x)$ under convergence in probability which, although a weaker form of
convergence, leads to the same limit as that for a.s.\ convergence.
We therefore need to show that, for any $\epsilon>0$, $\lim_{k\rightarrow \infty} \pr[|V_k(x) -V^*(x)|>\epsilon]=0$.
Denoting ${\hat V}(x)=\lim_{k \rightarrow \infty} V_k(x)$ and defining
$${\hat Q}(x, a) := \limsup_{k\rightarrow\infty} Q_k(x,a) = \limsup_{k\rightarrow\infty} \RobustOp Q_k(x,a),$$
it is readily apparent that we simply need to show
\begin{equation} \label{eqn:bd}
\lim_{k\rightarrow \infty} \pr[ {\hat Q}(x, a) - \BellmanOp {\hat Q}(x, \pi^{b_k})>\epsilon]=0 \quad \mbox{and} \quad
\lim_{k\rightarrow \infty} \pr[ {\hat Q}(x, a) - \BellmanOp {\hat Q}(x, \pi^{b_k})<-\epsilon]=0,
\end{equation}
since
\eqref{eqn:bd} leads
to ${\hat V}(x) = \max_a \{ R(x, a) \} + \gamma \ex[{\hat V}(x')]$,
which is the equation that is uniquely satisfied by $V^*(x)$.

To this end, we first observe that the event
$$\{ Q_{k}(x,a) \le \BellmanOp Q_k(x,a)+\epsilon\}$$
happens infinitely often in the limit as $k$ goes to infinity,
due to the a.s.\ convergence established above and the fact that $Q_{k+1}(x,a) \le \BellmanOp Q_k(x,a)$ is always true for any $k$ because of the
ordering by definition between the robust stochastic operator and the Bellman operator.
This then implies \eqref{eqn:bd} (left).
At the same time, we have
$$Q_{k+1}(x,a)  \; = \; \BellmanOp Q_k - \beta_k[V_k(x) - Q_k(x,a)],$$
which leads to the observation that the event 
\begin{align*}
\{Q_k(x,a) \ge \BellmanOp Q_k -\overline{\beta}_k[V_k(x) - Q_k(x,a)]- \epsilon\}
\end{align*}
happens infinitely often in the limit as $k$ goes to infinity.
This, together with the fact that $\overline{\beta}_k\le 1$, implies \eqref{eqn:bd} (right).
Hence, the desired relationship ${\hat V}(x)=V^*(x)$ holds a.s., which establishes the preservation of optimality.

Now, turning to prove that $\RobustOp$ is gap-increasing in a stochastic sense, the above arguments render $\lim_{k\rightarrow\infty} V_k(x) = V^*(x)$ a.s.,
and thus \eqref{eq:gap-inc} is equivalent to $\limsup_{k\rightarrow\infty} Q_k(x,a) \leq Q^*(x,a)$ a.s.
This inequality follows on a sample path basis from $\RobustOp Q(x,a) \leq \BellmanOp Q(x,a)$ by definition and our above arguments,
and thus we have the desired result for the operators $\RobustOp$.
Furthermore, it is readily verified that the above arguments can be similarly applied to cover all of the operators in $\RobustOpFamily$.

Lastly, from the above results of \eqref{eq:gap-inc} and ${\hat V}(x) = V^*(x)$ a.s.,
it follows that~\eqref{eq:opt-pre} also holds a.s.\ for $\RobustOp$ as well as for all operators in $\RobustOpFamily$, thus completing the proof.

\subsection{Proof of Theorem~\ref{thm:var1}}
%
First, we prove that ${\hat Q}_k \le_{st} {\tilde Q}_k$ holds for every $k\ge 1$, arguing by induction.
Suppose that this holds true for certain $k$, then the identity 
\begin{align*}
Q_{k+1}(x,a) = \rewardR(x,a) + \gamma \ex_{\pr} \max_{b \in \setA} Q_k(x',b) - \beta_k (V_k(x) - Q_k(x,a)),
\end{align*}
together with the fact that $\hat{\beta}_k\ge_{st}\tilde{\beta}_k$, yields
\begin{align*}
\ex[f(\hat{Q}_{k+1})|\hat{Q}_{k}] \le \ex[f(\tilde{Q}_{k+1})|\tilde{Q}_{k}]
\end{align*}
for any increasing function $f(\cdot)$, as long as the expectations exist.
Furthermore, by the induction assumption, we can conclude that $\ex[f(\hat{Q}_{k+1})] \le \ex[f(\tilde{Q}_{k+1})]$, and therefore ${\hat Q}_k \le_{st} {\tilde Q}_k$. 
Meanwhile, for any state-action pair $(x, a)$ in these systems, the action gap is characterized by the quantity
\begin{align*}
\lim \inf_{k \rightarrow \infty} V_k(x) - Q_k(x,a).
\end{align*}
Equivalently, we have
\begin{align*}
V^*(x) - \lim \sup_{k \rightarrow \infty}  Q_k(x,a),
\end{align*}
since we know that for both sequences $\{\hat{\beta}_k\}$ and $\{\tilde{\beta}_k\}$, the robust stochastic operator is optimality preserving.
We therefore obtain
\begin{align*}
\ex[f(V^*(x) -\lim \sup_{k \rightarrow \infty} \hat{Q}_{k+1})] \ge \ex[f( V^*(x)-\lim \sup_{k \rightarrow \infty} \tilde{Q}_{k+1})]
\end{align*}
for any increasing function, which follows from the fact that the limit preserves the stochastic order.
Hence, the stochastic order of the action gap is established.

\subsection{Proof of Theorem~\ref{thm:var2}}
%
We follow along similar lines for the proof of Theorem~\ref{thm:var1}.
With $f(x)$ being a convex function (and so is $f(-x)$) and with the identity
\begin{align*}
Q_{k+1}(x,a) = \rewardR(x,a) + \gamma \ex_{\pr} \max_{b \in \setA} Q_k(x',b) - \beta_k (V_k(x) - Q_k(x,a)),
\end{align*}
we can prove by induction that ${\hat Q}_k \ge_{cx} {\tilde Q}_k$.
Then, for any convex function $f(\cdot)$, we have
\begin{align*}
\ex[f(V^*(x) -\lim \sup_{k \rightarrow \infty} \hat{Q}_{k+1})] \ge \ex[f( V^*(x)-\lim \sup_{k \rightarrow \infty} \tilde{Q}_{k+1})],
\end{align*}
and thus establishing the convex order of the action gaps. 

\subsection{Proof of Theorem~\ref{thm:var3}}
%
The desired result can be readily seen from
\begin{align*}
\var[\hat{Q}_{k+1}] &= \ex[\var[\hat{Q}_{k+1}|\hat{Q}_k]+ \var[\ex[\hat{Q}_{k+1}|\hat{Q}_k]] \\ & = \var[\hat{\beta}_k]\ex[(\hat{V}_k(x) - \hat{Q}_k(x,a))^2] +  \var[\ex[\hat{Q}_{k+1}|\hat{Q}_k]]
\end{align*}
and
\begin{align*}
\var[\tilde{Q}_{k+1}] &= \ex[\var[\tilde{Q}_{k+1}|\tilde{Q}_k]+ \var[\ex[\tilde{Q}_{k+1}|\tilde{Q}_k]] \\ & = \var[\tilde{\beta}_k]\ex[(\tilde{V}_k(x) - \tilde{Q}_k(x,a))^2] +  \var[\ex[\tilde{Q}_{k+1}|\tilde{Q}_k]] .
\end{align*}

\clearpage
\newpage
\newpage
\newpage

\section{Python Code}
\label{app:code}
We tested the various operators of interest on several RL problems and algorithms.
For our empirical comparisons, the existing code that updates the Q-learning value based on the Bellman operator $\BellmanOp$
is replaced with the corresponding code for the $\ConsistentOp$ and $\RobustOp$ operators.
In particular, the snippets of code in Figure~\ref{fig:py-code} describe how this is generically implemented for the original
$\BellmanOp$ operator together with the added $\ConsistentOp$ and $\RobustOp$ operators, respectively.

\begin{figure*}[htbp]
\begin{verbatim}
def UpdateQBellman(self,currentState,action,nextState,reward,alpha,gamma):
  Qvalue=self.Q[currentState,action]
  rvalue=reward+gamma*max([self.Q[nextState,a] for a in self.actionsSet])
  self.Q[currentState,action] += alpha*(rvalue - Qvalue)

def UpdateQConsistent(self,currentState,action,nextState,reward,alpha,gamma):
  Qvalue=self.Q[currentState,action]
  rvalue=reward+gamma*(max([self.Q[nextState,a] for a in self.actionsSet]) 
      if currentState != nextState else Qvalue) 
  self.Q[currentState,action] += alpha*(rvalue - Qvalue)
	
def UpdateQRSO(self,currentState,action,nextState,reward,alpha,gamma,beta):
  Qvalue=self.Q[currentState,action]
  rvalue=reward+(gamma*(max([self.Q[nextState,a] for a in self.actionsSet]))
      -beta*(max([self.Q[currentState,a] for a in self.actionsSet])-Qvalue))
  self.Q[currentState,action] += alpha*(rvalue - Qvalue)
\end{verbatim}
\caption{Generic python code for all three operators}
\label{fig:py-code}
\end{figure*}

\end{document}